\PassOptionsToPackage{table}{xcolor}
\documentclass[sigconf,nonacm]{acmart}
\AtBeginDocument{%
  }

\usepackage{xcolor}
\usepackage{booktabs}
\usepackage{makecell}
\usepackage{multirow}
\usepackage{pifont}

\begin{document}

\title{S$^3$Geo: Structure-Semantic Synergistic Learning for Cross-View Geo-Localization}

\author{Ziqian Mo}
\orcid{0009-0002-4943-9914}
\affiliation{%
  \institution{The Hong Kong University of Science and Technology (Guangzhou)}
  \city{Guangzhou}
  \country{China}}
\email{ziqianmo2023@163.com}

\author{Hill Zhang}
\orcid{0009-0002-0597-5301}
\affiliation{%
  \institution{Claremont McKenna College}
  \city{Claremont}
  \state{California}
  \country{USA}}
\email{yzhang68@students.claremontmckenna.edu}

\author{Haosheng Tan}
\orcid{0009-0004-7988-962X}
\affiliation{%
  \institution{The Hong Kong University of Science and Technology (Guangzhou)}
  \city{Guangzhou}
  \country{China}}
\email{su21822@alumni.bristol.ac.uk}

\author{Ling Li}
\orcid{0009-0003-6899-1314}
\affiliation{%
  \institution{The Hong Kong University of Science and Technology (Guangzhou)}
  \city{Guangzhou}
  \country{China}}
\email{lli297@connect.hkust-gz.edu.cn}

\author{Jiaheng Wei}
\orcid{0009-0008-4950-4994}
\authornote{Corresponding author}
\affiliation{%
  \institution{The Hong Kong University of Science and Technology (Guangzhou)}
  \city{Guangzhou}
  \country{China}}
\email{jiahengwei@hkust-gz.edu.cn}

\renewcommand{\shortauthors}{Mo et al.}

\begin{abstract}
  Cross-view geo-localization (CVGL) aims to estimate geographic locations by matching images captured from different viewpoints, such as drone and satellite views. Existing methods mainly rely on visual representations, but often fail to jointly model fine-grained structural correspondences and semantic priors, making them prone to confusion between visually similar but semantically different regions, and thus limiting robustness under large viewpoint variations. To address these challenges, we propose \textbf{S$^3$Geo}, a structure-semantic synergistic learning framework for cross-view matching. Specifically, we first introduce a Decoupled Query Pooling (DQP) module to extract a compact set of region-aware features from dense tokens, enabling explicit modeling of local structural patterns. We then design a query-level contrastive learning scheme with an optimal transport (OT)-based formulation to establish soft correspondences under cross-view spatial misalignment. Furthermore, we incorporate a Semantic Knowledge Distillation (SKD) strategy from a frozen CLIP teacher to transfer semantic priors and relational structures, thereby improving discrimination on hard negatives. By operating synergistically, the semantic priors provide robust contextual filtering, which guides the structural module to establish precise spatial alignments. Experiments on the University-1652 and SUES-200 datasets demonstrate that \textbf{S$^3$Geo} consistently outperforms state-of-the-art approaches without increasing inference complexity, validating the effectiveness of jointly modeling structural and semantic information for CVGL.
\end{abstract}



\maketitle

\section{Introduction}

Cross-view geo-localization (CVGL) is commonly formulated as an image retrieval task across heterogeneous platforms, where the goal is to estimate geographic locations of a query image by matching it with reference images captured from different viewpoints, such as unmanned aerial vehicles (UAVs) and satellites~\cite{zheng2020university, zhu2023sues, li2026survey}.
The task was originally introduced for ground-to-satellite matching, providing a visual alternative to Global Navigation Satellite Systems (GNSS) \cite{tian2017cross} in urban environments where severe occlusions from high-rise buildings degrade positioning accuracy. With the rapid development of UAV technologies, UAV imagery has been incorporated into CVGL, enabling large-scale spatial observation and alleviating viewpoint discrepancies \cite{chowdhury2017drones, miranda2022autonomous}. Compared with ground-level images, UAV images offer wider coverage, fewer occlusions, and viewpoints more geometrically consistent with satellite imagery, making them suitable for cross-view matching. In practical scenarios, UAV-to-satellite CVGL plays a crucial role in UAV localization and navigation, supporting accurate positioning and path planning in applications such as disaster response \cite{chowdhury2017drones, nex2022uav}, environmental monitoring \cite{potter2019environmental, gaffey2020applications}, and infrastructure inspection \cite{besada2018drone, he2024uav}.

\begin{figure}[h]
	\centering
  \vspace{-0.1in}
	\includegraphics[width=0.48\textwidth]{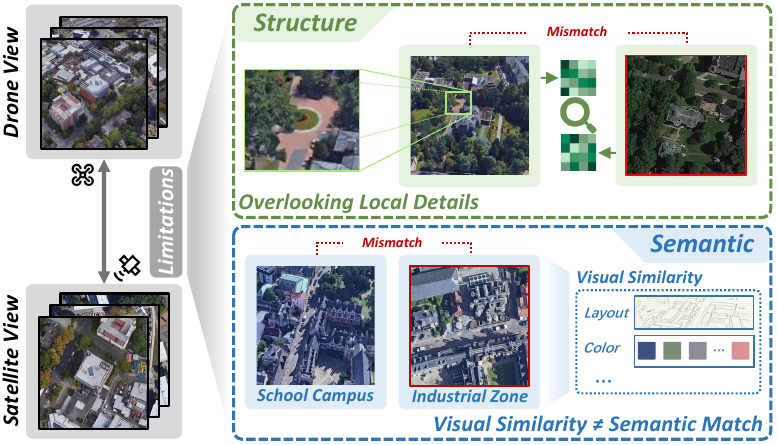} 
	\caption{Illustration of structural and semantic limitations in CVGL. Global representations may overlook local structural differences, while visually similar regions may correspond to different semantic categories, causing mismatches.}
  \Description{Illustration of structural and semantic challenges in cross-view geo-localization, showing mismatches caused by local structural differences and semantic ambiguity.}
	\label{fig:intro}
\end{figure}

Despite the significant progress achieved by recent deep learning-based methods \cite{wang2021each, dai2021transformer, deuser2023sample4geo, chen2025multi}, most existing approaches rely on a single visual representation for matching, leading to several inherent limitations. As illustrated in Figure ~\ref{fig:intro}, global representations may overlook fine-grained structural differences under similar layouts, while visually similar regions may correspond to different semantic categories, causing mismatches.
From the perspective of structural modeling, many methods perform alignment based on global features \cite{zheng2020university, zhu2022transgeo}, which tend to overlook discriminative local regions and fail to establish fine-grained spatial correspondences. Although some recent works attempt to incorporate local feature modeling or attention mechanisms to capture structural information, they often rely on implicit alignment or assume consistent spatial layouts across views, limiting their effectiveness under severe viewpoint variations \cite{wang2021each, shi2019spatial}. From the perspective of semantic modeling, purely visual models lack semantic priors, making them prone to mismatches when encountering scenes with similar structures but different semantic meanings. Recent efforts introduce semantic cues or vision-language models to enhance representation learning~\cite{vivanco2023geoclip, gao2025semantic, chen2025cross, li2024georeasoner, li2026recognition}. However, these approaches typically focus on feature-level enhancement and lack modeling of the relational structure of semantic knowledge, thereby limiting their discriminative capability in cross-view matching. Meanwhile, severe viewpoint variations lead to inherent cross-view misalignment, where local regions often do not follow strict one-to-one correspondences. Therefore, jointly modeling local structural alignment and semantic knowledge, while enabling flexible correspondences, remains a critical challenge for CVGL.

To address these challenges, we propose \textbf{S$^3$Geo}, a structure-semantic synergistic learning framework for CVGL. To explicitly model local structures and mitigate spatial misalignment, we introduce a Decoupled Query Pooling (DQP) module equipped with an Optimal Transport (OT)-based query-level contrastive loss \cite{cuturi2013sinkhorn, sarlin2020superglue} for soft cross-view matching. To resolve semantic ambiguity, we design a Semantic Knowledge Distillation (SKD) strategy that transfers language-aligned priors from CLIP \cite{radford2021learning}. Crucially, this distilled knowledge acts as a macro-level filter to reject visually deceptive hard negatives, enabling DQP to establish precise micro-level spatial correspondences. Without extra inference overhead, \textbf{S$^3$Geo} unifies structural and semantic learning, achieving state-of-the-art performance on the University-1652 \cite{zheng2020university} and SUES-200 \cite{zhu2023sues} benchmarks. Our contributions are summarized as follows:
\begin{itemize}
    \item We propose \textbf{S$^3$Geo}, a structure-semantic synergistic learning framework that unifies local structural modeling and semantic knowledge enhancement for CVGL.
    \item We design two complementary mechanisms: a Decoupled Query Pooling (DQP) module with OT-based soft matching for flexible structural alignment, and a Semantic Knowledge Distillation (SKD) strategy for transferring language-aligned semantic priors and relational structures without additional annotations.
    \item Extensive experiments on public benchmarks demonstrate that our method consistently outperforms existing approaches, validating the effectiveness of the proposed framework.
\end{itemize}

\section{Related Work}

\subsection{Cross-view Geo-localization}

Early research in CVGL primarily relies on convolutional neural networks (CNNs) to extract global visual representations and perform retrieval via metric learning \cite{lin2015learning, hu2018cvm, zheng2020university}. While effective to some extent, these methods often struggle with severe viewpoint variations due to limited modeling of long-range dependencies and spatial context.

More recent works adopt transformer-based \cite{vaswani2017attention} architectures to enhance feature representation by capturing global context and spatial relationships \cite{dai2021transformer, zhao2024transfg, li2024geoformer}. Despite these advances, most existing approaches still rely on global representations \cite{chen2024sdpl, du2024ccr, zhou2025cdm}, which may overlook fine-grained structural details and remain sensitive to cross-view misalignment. These limitations highlight the need for more effective modeling of region-level correspondences and semantic consistency.

\subsection{Structural Modeling for Cross-view Matching}

To alleviate the spatial misalignment caused by viewpoint discrepancies, several studies attempt to incorporate local feature modeling or attention-based mechanisms. Patch-based representations, multi-branch architectures, and transformer attention have been explored to capture spatial structures and improve matching robustness \cite{deuser2023sample4geo, shen2023mccg, ge2024multibranch}. Some methods further exploit multi-scale features or cross-attention pooling strategies to enhance structural representation \cite{wu2024camp}. Additionally, spatial aggregation mechanisms, such as graph convolutional networks and direction-aware learning, have been introduced to explicitly model geometric topologies \cite{liu2024segcn, lv2024direction}. 

Nevertheless, these approaches often rely on implicit alignment or assume consistent spatial layouts across views, which is not valid in cross-view scenarios. As a result, they struggle to establish reliable fine-grained correspondences between regions under drastic viewpoint changes. Moreover, naive one-to-one matching of local features may introduce incorrect associations due to inconsistent region ordering. These limitations highlight the need for more flexible region-level representations and correspondence modeling strategies that can adapt to cross-view misalignment without relying on strict spatial assumptions.

\subsection{Semantic Learning for Cross-view Matching}

Most existing highly competitive CVGL methods \cite{xia2024enhancing, chen2025multi, fang2025scof, yang2026integrating, mo2026star} still focus primarily on visual representation learning and lack explicit semantic modeling, making them vulnerable to semantic ambiguity and hard negative samples. In many cases, visually similar regions may correspond to different semantic categories, leading to incorrect matching results. Recent advances in vision-language models, such as CLIP \cite{radford2021learning}, provide language-aligned semantic representations that capture high-level concepts beyond purely visual features. These models have been widely applied in various vision tasks to improve semantic understanding and generalization. 

Some recent works, such as SIGN \cite{mo2025sign}, incorporate semantic cues to enhance representation learning. CLIP-UG \cite{wu2025clip} further leverages CLIP to introduce language-aligned semantic features into the matching framework. However, it mainly relies on feature-level fusion and lacks an explicit mechanism to effectively transfer semantic knowledge while preserving the relational structure of cross-view similarities. As a result, effectively transferring semantic knowledge for cross-view matching remains underexplored, especially for improving the discrimination of hard negative samples under complex scene variations.

\begin{figure*}[t]
	\centering
	\includegraphics[width=\textwidth]{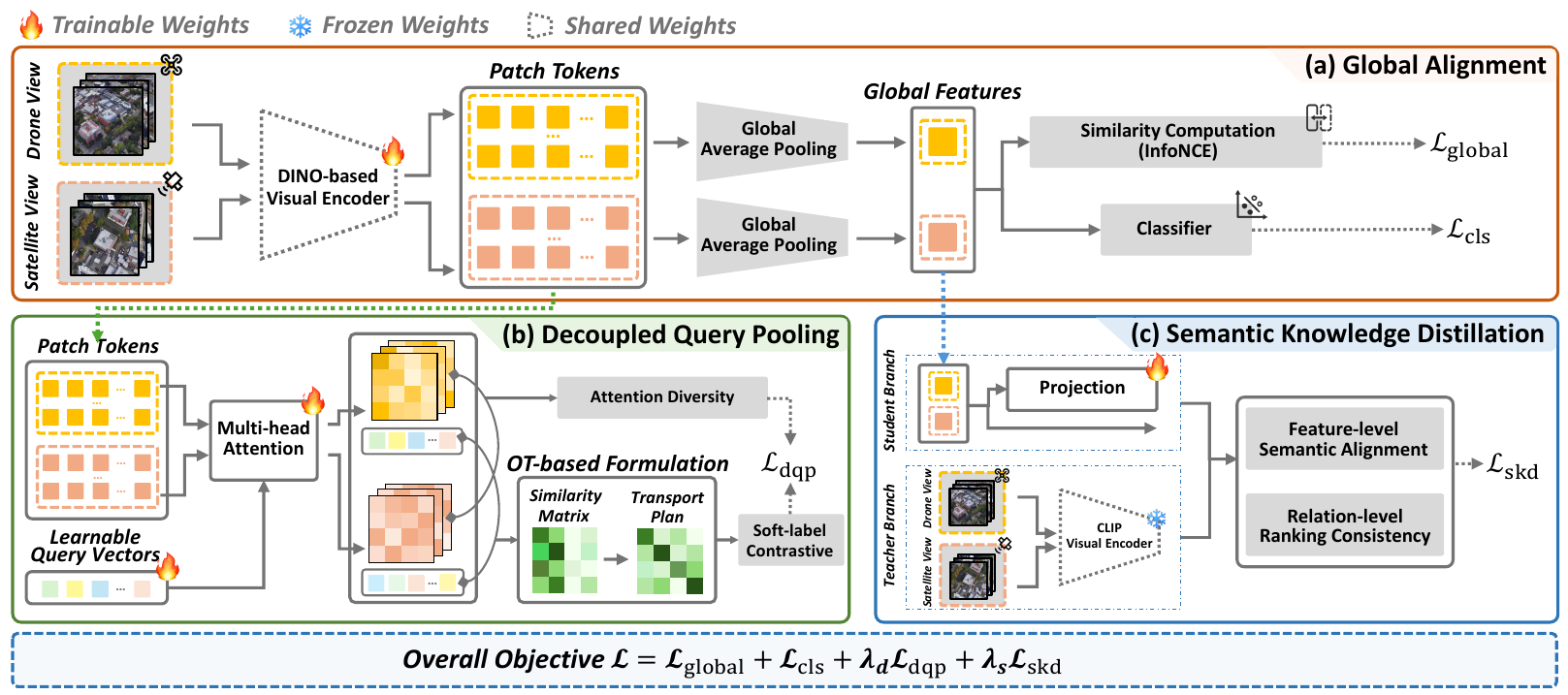} 
	\caption{Overview of \textbf{S$^3$Geo}. (a) Global alignment learns instance-level representations via contrastive learning and classification. (b) The DQP module performs region-level structural modeling with OT-based matching with soft-label contrastive learning and attention diversity. (c) The SKD module enhances semantic consistency via feature- and relation-level distillation from CLIP. During inference, only the global branch is used for retrieval, while DQP and SKD are employed for training without additional overhead.}
  \Description{Overview of the S$^3$Geo framework, including global alignment, decoupled query pooling, semantic knowledge distillation, and the overall training objective.}
	\label{fig:framework}
    \vspace{-0.5cm}
\end{figure*}

\section{Method}

We propose \textbf{S$^3$Geo}, a structure-semantic synergistic learning framework for CVGL. As illustrated in Figure ~\ref{fig:framework}, our method jointly models global alignment, local structural correspondence, and semantic consistency to address cross-view misalignment and semantic ambiguity. Given a pair of images from different viewpoints (e.g., drone and satellite), we extract visual features using a shared backbone and perform cross-view matching based on the learned representations.

We first introduce the backbone and global alignment, followed by the DQP module for structural modeling and the SKD strategy for semantic enhancement. Finally, we present the overall training objective.

\subsection{Backbone Representation and Global Alignment}

Given a drone image $x^q$ and a satellite image $x^g$, we adopt a DINO-based visual encoder \cite{simeoni2025dinov3} $f_{\text{dino}}(\cdot)$ to extract dense visual representations in the form of patch tokens:
\begin{equation}
\mathbf{Z}^q = f_{\text{dino}}(x^q) \in \mathbb{R}^{N \times D}, \qquad
\mathbf{Z}^g = f_{\text{dino}}(x^g) \in \mathbb{R}^{N \times D},
\end{equation}
where $N$ denotes the number of visual tokens and $D$ is the feature dimension. The shared encoder ensures that features from different viewpoints are projected into a unified embedding space, providing a consistent basis for cross-view alignment.

To obtain compact image-level representations, we perform global average pooling over all tokens:
\begin{equation}
\mathbf{v}^q = \frac{1}{N}\sum_{n=1}^{N}\mathbf{Z}_n^q, \qquad
\mathbf{v}^g = \frac{1}{N}\sum_{n=1}^{N}\mathbf{Z}_n^g.
\end{equation}

These global features capture holistic scene information and serve as the foundation for instance-level matching.

To establish cross-view alignment, we adopt a bidirectional InfoNCE objective \cite{oord2018representation}. After $L_2$ normalization, the similarity between drone and satellite features is computed as:
\begin{equation}
\mathbf{M}_{ij} = \tau^{-1}\,\hat{\mathbf{v}}_i^q \cdot \hat{\mathbf{v}}_j^g,
\end{equation}
where $\hat{\mathbf{v}}=\mathbf{v}/\|\mathbf{v}\|_2$ and $\tau$ is a temperature parameter controlling the concentration level of the similarity distribution. The global contrastive loss is formulated as:
\begin{equation}
\mathcal{L}_{\mathrm{global}} = \frac{1}{2B}\left[
\sum_{i=1}^{B}\mathrm{CE}(\mathbf{S}_{i,:},i) + \sum_{j=1}^{B}\mathrm{CE}(\mathbf{S}_{:,j},j) \right],
\end{equation}
where $B$ is the batch size. This bidirectional formulation enforces mutual consistency between drone$\rightarrow$satellite and satellite$\rightarrow$drone retrieval, encouraging matched pairs to be close in the embedding space while pushing apart unmatched pairs.

In addition to contrastive supervision, we introduce an identity (ID) classification loss to further enhance feature discriminability. Specifically, we treat each geographic location as a distinct class and train a classifier $\phi(\cdot)$ on top of the global representation:

\begin{equation}
\mathbf{p}_i^q = \phi(\mathbf{v}_i^q), \qquad
\mathbf{p}_i^g = \phi(\mathbf{v}_i^g),
\end{equation}
\begin{equation}
\mathcal{L}_{\mathrm{cls}} = \frac{1}{2B}\sum_{i=1}^{B} \left[\mathrm{CE} (\mathbf{p}_i^q,y_i)+ \mathrm{CE}(\mathbf{p}_i^g,y_i) \right],
\end{equation}
where $y_i$ denotes the ground-truth location label. This formulation introduces category-level supervision, which complements instance-level contrastive learning by improving inter-class separability and stabilizing optimization.

Together, $\mathcal{L}_{\mathrm{global}}$ and $\mathcal{L}_{\mathrm{cls}}$ form the basic training objective for global cross-view retrieval.

\subsection{Decoupled Query Pooling Module}
While global contrastive learning enforces instance-level alignment, it does not explicitly model fine-grained spatial correspondences across views. In particular, global pooling aggregates all patch tokens into a single representation, inevitably suppressing discriminative local structures such as road layouts and building configurations, which are crucial for cross-view matching.

To capture such local structures, we employ a DQP module to extract a compact set of region-level features from dense patch tokens. Specifically, we define $M$ learnable query vectors
\(\mathbf{Q}_0 \in \mathbb{R}^{M \times D},\)
which are randomly initialized and optimized jointly with the network. These learnable queries interact with patch tokens through multi-head cross-attention \cite{vaswani2017attention}:
\begin{equation}
\mathbf{Q}^q, \mathbf{A}^q = \mathrm{MHA}(\mathbf{Q}_0,\mathbf{Z}^q,\mathbf{Z}^q), \qquad
\mathbf{Q}^g, \mathbf{A}^g = \mathrm{MHA}(\mathbf{Q}_0,\mathbf{Z}^g,\mathbf{Z}^g),
\end{equation}
where $\mathbf{Q}^q,\mathbf{Q}^g\in\mathbb{R}^{M\times D}$ denote query-level features, and $\mathbf{A}^q,\mathbf{A}^g\in\mathbb{R}^{M\times N}$ are the corresponding attention maps. Unlike heuristic saliency-based selection or fixed region partitioning, DQP performs data-driven structural decomposition, where each query acts as a region probe that selectively aggregates informative tokens, producing complementary regional descriptors.

While DQP provides compact region-aware representations, it does not explicitly resolve cross-view correspondence. In particular, the extracted query features lack consistent spatial ordering across views, making one-to-one alignment unreliable under significant viewpoint variations. To address this, we adopt an Optimal Transport (OT)-based formulation to model soft correspondences between query features~\cite{cuturi2013sinkhorn, sarlin2020superglue}. For each positive pair $(x_i^q,x_i^g)$, we compute the pairwise similarity matrix as:
\begin{equation}
\mathbf{K}_i = \hat{\mathbf{Q}}_i^q(\hat{\mathbf{Q}}_i^g)^\top \in \mathbb{R}^{M\times M},
\end{equation}
where $\hat{\mathbf{Q}}$ denotes $L_2$-normalized query features, and define the transport cost as:
\begin{equation}
\mathbf{C}_i=\mathbf{1}_{M\times M}-\mathbf{K}_i.
\end{equation}

We then solve the entropy-regularized OT problem as:
\begin{equation}
\mathbf{P}_i = \arg\min_{\mathbf{P}\in\Pi(\mathbf{a},\mathbf{b})} \langle \mathbf{P},\mathbf{C}_i\rangle -\varepsilon H(\mathbf{P}),
\end{equation}
where $\mathbf{a}=\mathbf{b}=\frac{1}{M}\mathbf{1}$ denote uniform marginals, $\Pi(\mathbf{a},\mathbf{b})$ denotes the set of admissible transport plans with marginals $\mathbf{a}$ and $\mathbf{b}$, and $H(\mathbf{P})$ is the entropy regularization term.

To jointly train all query tokens, we flatten them across the batch and compute cross-view logits:
\begin{equation}
\mathbf{Z}_{(i,m),(j,n)} = \tau_u^{-1}\, \hat{\mathbf{Q}}_{i,m}^q \cdot \hat{\mathbf{Q}}_{j,n}^g,
\end{equation}
where $m,n \in \{1,\dots,M\}$ index the query tokens, and $\tau_u$ is a temperature parameter. This formulation treats each query token as an individual matching unit, allowing all region-level correspondences within a mini-batch to be optimized jointly. We construct the corresponding soft label matrix as:
\begin{equation}
\mathbf{Y}_{(i,m),(i,n)}=\mathbf{P}_i(m,n), \qquad
\mathbf{Y}_{(i,m),(j,n)}=0 \ (i\neq j).
\end{equation}

Finally, we define a symmetric soft-label contrastive loss with an additional decorrelation regularization:

\begin{equation}
\begin{aligned}
\mathcal{L}_{\mathrm{dqp}} =\;& 
\frac{1}{2}\Big(\mathrm{CE}_{\mathrm{soft}}(\mathbf{Z},\mathbf{Y}) + \mathrm{CE}_{\mathrm{soft}}(\mathbf{Z}^{\top},\mathbf{Y}^{\top})\Big) \\
&+ \alpha \Big(
\left\| \hat{\mathbf{A}}^q(\hat{\mathbf{A}}^q)^{\top}-\mathbf{I}_M \right\|_F^2 
+ \left\| \hat{\mathbf{A}}^g(\hat{\mathbf{A}}^g)^{\top}-\mathbf{I}_M \right\|_F^2
\Big),
\end{aligned}
\end{equation}
which enforces consistent region-level alignment across views and complements global contrastive learning with fine-grained structural supervision. Note that the second term applies a decorrelation regularization to the attention maps of both views independently, where $\hat{\mathbf{A}}$ denotes the row-wise $L_2$-normalized attention maps along the spatial dimension. Meanwhile, the scaling constant $\alpha$ controls this penalty, encouraging different queries to attend to complementary spatial regions in both drone and satellite images, thereby reducing redundancy among the learned region-level representations.

\subsection{Semantic Knowledge Distillation Strategy}

In cross-view scenarios, visually similar regions may correspond to semantically different locations, making it difficult to distinguish hard negatives for purely visual models. To address this limitation, instead of directly fusing CLIP features, we distill semantic knowledge from a frozen CLIP visual encoder \cite{radford2021learning, hinton2015distilling} $f_{\mathrm{clip}}(\cdot)$ into the retrieval model, which provides language-aligned semantic priors learned from large-scale image-text data.

The student global features are first projected into the CLIP embedding space through a learnable head $h(\cdot)$:
\begin{equation}
\hat{\mathbf{s}}_i^q=\frac{h(\mathbf{v}_i^q)}{\|h(\mathbf{v}_i^q)\|_2}, \qquad
\hat{\mathbf{s}}_i^g=\frac{h(\mathbf{v}_i^g)}{\|h(\mathbf{v}_i^g)\|_2},
\end{equation}
while the teacher features are computed as:
\begin{equation}
\hat{\mathbf{t}}_i^q = \frac{f_{\mathrm{clip}}(x_i^q)}{\|f_{\mathrm{clip}}(x_i^q)\|_2}, \qquad
\hat{\mathbf{t}}_i^g = \frac{f_{\mathrm{clip}}(x_i^g)}{\|f_{\mathrm{clip}}(x_i^g)\|_2}.
\end{equation}

This projection enables the student to align with the CLIP semantic space, thereby introducing semantic cues that are not captured by purely visual representations and improving robustness to appearance ambiguity. To further preserve the teacher's ranking behavior, we compute the cross-view similarity matrices:
\begin{equation}
\mathbf{R}_{ij}^{s} = \hat{\mathbf{v}}_i^q\cdot\hat{\mathbf{v}}_j^g, \qquad
\mathbf{R}_{ij}^{t} = \hat{\mathbf{t}}_i^q\cdot\hat{\mathbf{t}}_j^g,
\end{equation}
and transform them into probability distributions via row-wise softmax with temperature $T$:
\begin{equation}
\mathbf{P}_i^{s}=\mathrm{softmax}(\mathbf{R}_{i,:}^{s}/T), \qquad
\mathbf{P}_i^{t}=\mathrm{softmax}(\mathbf{R}_{i,:}^{t}/T).
\end{equation}

The student similarity is computed in the original retrieval space ($\hat{\mathbf{v}}$) rather than the projected space ($\hat{\mathbf{z}}$), ensuring that the distilled relational supervision directly benefits the target matching objective. Since cross-view retrieval is inherently a ranking problem, matching the similarity distributions allows the student to inherit the teacher's soft ranking structure, providing richer supervision for hard negatives.

The unified semantic knowledge distillation objective is defined as:
\begin{equation}
\mathcal{L}_{\mathrm{skd}} = \frac{1}{B}\sum_{i=1}^{B} \Big[ \big(1-\cos(\hat{\mathbf{s}}_i^q,\hat{\mathbf{t}}_i^q)\big) + \big(1-\cos(\hat{\mathbf{s}}_i^g,\hat{\mathbf{t}}_i^g)\big) + \beta\,\mathrm{KL}\big(\mathbf{P}_i^{t}\,\|\,\mathbf{P}_i^{s}\big)\Big].
\end{equation}

The KL divergence term preserves the relative similarity structure between samples, making it particularly suitable for transferring ranking information compared to point-wise regression objectives \cite{park2019relational}. The coefficient $\beta$ is a scaling factor to balance the magnitude of the KL divergence term for stable optimization.

This objective jointly enforces feature-level semantic alignment and relation-level ranking consistency. Ultimately, this semantic knowledge distillation complements our structural modeling, yielding more discriminative and robust cross-view representations.

\subsection{Overall Training Objective}

The overall training objective combines global alignment, classification supervision, local structural modeling, and semantic knowledge distillation. Specifically, we optimize:
\begin{equation}
\mathcal{L} = \mathcal{L}_{\mathrm{global}} + \mathcal{L}_{\mathrm{cls}} + \lambda_d \mathcal{L}_{\mathrm{dqp}} + \lambda_s \mathcal{L}_{\mathrm{skd}},
\end{equation}
where $\mathcal{L}_{\mathrm{global}}$ enforces instance-level alignment across views, $\mathcal{L}_{\mathrm{cls}}$ improves feature discriminability via identity supervision, $\mathcal{L}_{\mathrm{dqp}}$ provides fine-grained structural alignment at the region level, and $\mathcal{L}_{\mathrm{skd}}$ injects semantic knowledge from the CLIP teacher. During inference, only the global branch is retained for similarity computation, while the DQP and SKD modules are used as auxiliary supervision during training without introducing additional overhead.

\section{Experiments}

\subsection{Datasets and Evaluation Metrics}

We evaluate our method on two widely used CVGL benchmarks: University-1652 \cite{zheng2020university} and SUES-200 \cite{zhu2023sues}.

\textbf{University-1652} is a large-scale CVGL dataset containing drone-view, satellite-view, and ground-view images collected from 1,652 buildings across 72 universities. It is a standard benchmark for drone-to-satellite geo-localization. The dataset consists of 50,218 training images, 41,135 query images, and 55,227 gallery images. We follow the common evaluation settings of Drone $\to$ Satellite and Satellite $\to$ Drone.

\textbf{SUES-200} extends this setting by introducing multi-altitude drone imagery, with images captured at four flight heights (150m, 200m, 250m, and 300m). Compared with University-1652, it exhibits larger variations in scale and viewpoint, making it more challenging for fine-grained cross-view alignment.

\textbf{Evaluation Metrics.} Following prior works, we adopt Recall@K (R@K) and Average Precision (AP) for evaluation. R@K measures the percentage of queries for which at least one correct match appears in the top-K retrieved results, reflecting retrieval accuracy. AP summarizes the precision-recall trade-off over the entire ranking list, providing a more comprehensive evaluation of retrieval performance.

\subsection{Implementation Details}

We implement our method using PyTorch \cite{paszke2019pytorch} and conduct all experiments on a single NVIDIA A100 GPU. The input drone and satellite images are resized to $512 \times 512$. During training, we apply standard data augmentations, including random horizontal flipping, random cropping, and color jittering.

For the visual backbone, we adopt the DINOv3 encoder \cite{simeoni2025dinov3} to extract patch-level features. To preserve its robust generalized representations while adapting to the cross-view domain, we freeze all but the last two transformer blocks of the DINOv3 encoder during fine-tuning. For the semantic knowledge distillation, we employ the pre-trained CLIP ViT-L/14 visual encoder \cite{radford2021learning} as the frozen teacher. The model is trained under a standard training setting with a batch size of 32. We use the AdamW optimizer \cite{loshchilov2017decoupled} with an initial learning rate of $5\times10^{-4}$. A cosine learning rate scheduler \cite{loshchilov2016sgdr} is applied to dynamically adjust the learning rate, with the first 10\% of the training steps serving as a linear warm-up phase to stabilize optimization.

For the DQP module, the number of learnable queries is set to $M=4$. In the optimal transport formulation, we use an entropy regularization parameter $\varepsilon=0.1$ and perform Sinkhorn normalization \cite{cuturi2013sinkhorn} for efficient computation. The overall objective weights are empirically set to $\lambda_d=0.5$ and $\lambda_s=0.5$ to balance local structural modeling, and semantic knowledge transfer.

\begin{table}[t]
  \centering
  \caption{Performance comparison on the University-1652 dataset. The best results are highlighted in \textbf{bold}, and the second-best are \underline{underlined}.}
  \label{tab:university1652}
  \setlength{\tabcolsep}{2pt} 
  \renewcommand{\arraystretch}{1.1}
  \resizebox{\columnwidth}{!}{ 
  \begin{tabular}{l l cc cc}
  \toprule
  \multirow{2}{*}{\textbf{Method}} & \multirow{2}{*}{\textbf{Venue}} & \multicolumn{2}{c}{\textbf{Drone$\rightarrow$Satellite}} & \multicolumn{2}{c}{\textbf{Satellite$\rightarrow$Drone}} \\
  \cmidrule(lr){3-4} \cmidrule(lr){5-6}
   &  & \textbf{R@1} & \textbf{AP} & \textbf{R@1} & \textbf{AP} \\
  \midrule
  University-1652 \cite{zheng2020university} & ACM MM'20 & 58.49 & 63.13 & 71.18 & 58.74 \\
  FSRA \cite{dai2021transformer}        & TCSVT'22 & 82.25 & 84.82 & 87.87 & 81.53 \\
  TransFG \cite{zhao2024transfg}     & TGRS'24  & 84.01 & 86.31 & 90.16 & 84.61 \\
  MJRLIFS \cite{ge2024multibranch}     & TGRS'24  & 86.06 & 88.08 & 91.44 & 85.73 \\
  GeoFormer \cite{li2024geoformer}   & JSTARS'24& 89.08 & 90.83 & 92.30 & 88.54 \\
  SeGCN \cite{liu2024segcn}       & JSTARS'24& 89.18 & 90.89 & 94.29 & 89.65 \\
  MCCG \cite{shen2023mccg}        & TCSVT'23 & 89.64 & 91.32 & 94.30 & 89.39 \\
  SDPL \cite{chen2024sdpl}        & TCSVT'24 & 90.16 & 91.64 & 93.58 & 89.45 \\
  CCR  \cite{du2024ccr}        & TCSVT'24 & 92.54 & 93.78 & 95.15 & 91.80 \\
  Sample4Geo \cite{deuser2023sample4geo}   & ICCV'23  & 92.65 & 93.81 & 96.43 & 93.79 \\
  SRLN \cite{lv2024direction}        & TGRS'24  & 92.70 & 93.77 & 95.14 & 91.97 \\
  MEAN \cite{chen2025multi}        & TGRS'25  & 93.55 & 94.53 & 96.01 & 92.08 \\
  SCOF \cite{fang2025scof}        & TGRS'25  & 93.68 & 94.68 & 96.29 & 92.68 \\
  CAMP \cite{wu2024camp}        & TGRS'24  & 94.46 & 95.38 & 96.15 & 92.72 \\
  DAC \cite{xia2024enhancing}         & TCSVT'24 & 94.67 & 95.50 & 96.43 & 93.79 \\
  SIGN \cite{mo2025sign}        & IGARSS'25& 94.68 & 95.59 & 96.29 & \underline{94.33} \\
  ECSNet \cite{yang2026integrating}      & TGRS'26  & 94.80 & 95.68 & \underline{96.57} & 92.89 \\
  CDM-Net \cite{zhou2025cdm}     & TGRS'25  & \underline{95.13} & \underline{96.04} & 96.43 & 93.79 \\
  \midrule
  \rowcolor{gray!10} \textbf{S$^3$Geo (Ours)} & - & \textbf{96.19} & \textbf{96.86} & \textbf{96.72} & \textbf{95.10} \\
  \bottomrule
  \end{tabular}
  }
  \vspace{-0.5cm}
\end{table}

\begin{table*}[t]
  \centering
  \small
  \caption{Performance comparison on the SUES-200. The best results are highlighted in \textbf{bold}, and the second-best results are \underline{underlined}.}
  \label{tab:sues200}
  \renewcommand{\arraystretch}{0.85}
  \setlength{\tabcolsep}{10pt} 
  
  \begin{tabular}{l l cc cc cc cc}
    \toprule
    \multicolumn{10}{c}{\textbf{Drone$\rightarrow$Satellite}} \\
    \midrule
    \multirow{2}{*}{\textbf{Method}} &
    \multirow{2}{*}{\textbf{Publication}} &
    \multicolumn{2}{c}{150m} &
    \multicolumn{2}{c}{200m} &
    \multicolumn{2}{c}{250m} &
    \multicolumn{2}{c}{300m} \\ 
    \cmidrule(lr){3-4} \cmidrule(lr){5-6} \cmidrule(lr){7-8} \cmidrule(lr){9-10}
      & & R@1 & AP & R@1 & AP & R@1 & AP & R@1 & AP \\
    \midrule
    SUES-200 \cite{zhu2023sues} & TCSVT'23 & 55.65 & 61.92 & 66.78 & 71.55 & 72.00 & 76.43 & 74.05 & 78.26 \\
    MJRLIFS \cite{ge2024multibranch} & TGRS'24 & 77.57 & 81.30 & 89.50 & 91.40 & 92.58 & 94.21 & 97.40 & 97.92 \\
    FSRA \cite{dai2021transformer} & TCSVT'22 & 68.25 & 73.45 & 83.00 & 85.99 & 90.68 & 92.27 & 91.95 & 91.95 \\
    MBF \cite{zhu2023uav} & Sensors'23 & 85.62 & 88.21 & 87.43 & 90.02 & 90.65 & 92.53 & 92.12 & 93.63 \\
    SeGCN \cite{liu2024segcn} & JSTARS'24 & 90.80 & 92.32 & 91.93 & 93.41 & 92.53 & 93.90 & 93.33 & 94.61 \\
    SRLN \cite{lv2024direction} & TGRS'24 & 89.90 & 91.90 & 94.32 & 95.65 & 95.92 & 96.79 & 96.37 & 97.21 \\
    MCCG \cite{shen2023mccg} & TCSVT'23 & 82.22 & 85.47 & 89.38 & 91.41 & 93.82 & 95.04 & 95.07 & 96.20 \\
    SCOF \cite{fang2025scof} & TGRS'25 & 90.75 & 92.32 & 94.25 & 95.35 & 96.88 & 97.42 & 97.85 & 98.10 \\
    Sample4Geo \cite{deuser2023sample4geo} & ICCV'23 & 92.60 & 94.00 & 97.38 & 97.81 & 98.28 & 98.64 & 99.18 & 99.36 \\
    CDM-Net \cite{zhou2025cdm} & TGRS'25 & 93.78 & 95.16 & 97.62 & 98.16 & 98.28 & 98.69 & 99.20 & 99.31 \\
    CAMP \cite{wu2024camp} & TGRS'24 & 95.40 & 96.38 & 97.63 & 98.16 & 98.05 & 98.45 & 99.33 & 99.46 \\    
    MEAN \cite{chen2025multi} & TGRS'25 & 95.50 & 96.46 & 98.38 & 98.72 & \underline{98.95} & 99.17 & \underline{99.52} & \underline{99.63} \\
    SIGN \cite{mo2025sign} & IGARSS'25 & 96.50 & 97.35 & 97.93 & 98.33 & 98.70 & 98.83 & 99.44 & 99.54 \\
    DAC \cite{xia2024enhancing} & TCSVT'24 & 96.80 & 97.54 & 97.48 & 97.97 & 98.20 & 98.62 & 97.58 & 98.14 \\ 
    ECSNet \cite{yang2026integrating} & TGRS'26 & \underline{96.82} & \underline{97.60} & \underline{98.84} & \underline{98.80} & 98.92 & \underline{99.20} & 99.50 & 99.58 \\
    \midrule
    \rowcolor{gray!10} \textbf{S$^3$Geo (Ours)} & - & \textbf{99.23} & \textbf{99.41} & \textbf{99.63} & \textbf{99.72} & \textbf{99.98} & \textbf{99.98} & \textbf{100.00} & \textbf{100.00} \\ 
    \midrule
    \multicolumn{10}{c}{\textbf{Satellite$\rightarrow$Drone}} \\ 
    \midrule
    \multirow{2}{*}{\textbf{Method}} &
    \multirow{2}{*}{\textbf{Publication}} &
    \multicolumn{2}{c}{150m} &
    \multicolumn{2}{c}{200m} &
    \multicolumn{2}{c}{250m} &
    \multicolumn{2}{c}{300m} \\ 
    \cmidrule(lr){3-4} \cmidrule(lr){5-6} \cmidrule(lr){7-8} \cmidrule(lr){9-10}
      & & R@1 & AP & R@1 & AP & R@1 & AP & R@1 & AP \\
    \midrule
    SUES-200 \cite{zhu2023sues} & TCSVT'23 & 75.00 & 55.46 & 85.00 & 66.05 & 86.25 & 69.94 & 88.75 & 74.46 \\
    MJRLIFS \cite{ge2024multibranch} & TGRS'24 & 93.75 & 79.49 & 97.50 & 90.52 & 97.50 & 96.03 & \textbf{100.00} & 97.66 \\
    FSRA \cite{dai2021transformer} & TCSVT'22 & 83.75 & 76.67 & 90.00 & 85.34 & 93.75 & 90.17 & 95.00 & 92.03 \\
    MBF \cite{zhu2023uav} & Sensors'23 & 88.75 & 84.74 & 91.25 & 89.95 & 93.75 & 90.65 & 96.25 & 91.60 \\
    SeGCN \cite{liu2024segcn} & JSTARS'24 & 93.75 & 92.45 & 95.00 & 93.65 & 96.25 & 94.39 & 97.50 & 94.55 \\
    SRLN \cite{lv2024direction} & TGRS'24 & 93.75 & 93.01 & 97.50 & 95.08 & 97.50 & 96.52 & 97.50 & 96.71 \\
    MCCG \cite{shen2023mccg} & TCSVT'23 & 93.75 & 89.72 & 93.75 & 92.21 & 96.25 & 96.14 & 98.75 & 96.64 \\
    SCOF \cite{fang2025scof} & TGRS'25 & 95.00 & 89.72 & 97.50 & 93.13 & 98.75 & 96.33 & 97.50 & 96.62 \\
    Sample4Geo \cite{deuser2023sample4geo} & ICCV'23 & 97.50 & 93.63 & 98.75 & 96.70 & 98.75 & 98.28 & 98.75 & 98.05 \\
    CDM-Net \cite{zhou2025cdm} & TGRS'25 & 95.25 & 92.24 & 98.50 & 96.40 & 99.00 & 97.60 & 99.00 & 98.01 \\
    CAMP \cite{wu2024camp} & TGRS'24 & 96.25 & 93.69 & 97.50 & 96.76 & 98.75 & 98.10 & \textbf{100.00} & 98.85 \\
    MEAN \cite{chen2025multi} & TGRS'25 & 97.50 & 94.75 & \textbf{100.00} & 97.09 & \textbf{100.00} & 98.28 & \textbf{100.00} & 99.21 \\
    SIGN \cite{mo2025sign} & IGARSS'25 & 98.75 & 95.67 & 98.75 & 97.50 & 98.75 & 98.28 & \textbf{100.00} & 98.86 \\
    DAC \cite{xia2024enhancing} & TCSVT'24 & 97.50 & 94.06 & 98.75 & 96.66 & 98.75 & 98.09 & 98.75 & 97.87 \\
    ECSNet \cite{yang2026integrating} & TGRS'26 & \underline{97.52} & \underline{95.01} & \underline{99.26} & \underline{97.20} & \underline{99.48} & \underline{98.30} & \underline{99.92} & \underline{99.30} \\
    \midrule
    \rowcolor{gray!10} \textbf{S$^3$Geo (Ours)} & - & \textbf{100.00} & \textbf{97.87} & \textbf{100.00} & \textbf{98.62} & \textbf{100.00} & \textbf{98.72} & \textbf{100.00} & \textbf{99.35} \\
    \bottomrule
  \end{tabular}
  \vspace{-0.3cm}
\end{table*}

\subsection{Comparison with State-of-the-Art Methods}

We compare the proposed \textbf{S$^3$Geo} with a broad range of recent state-of-the-art CVGL methods, including representative metric-learning, graph-based, and transformer-inspired approaches, as well as recent strong baselines built on modern visual backbones. The compared methods cover both early CVGL benchmarks and the latest high-performance models, providing a comprehensive evaluation of the proposed method. Quantitative results on the University-1652 and SUES-200 datasets are reported in Table~\ref{tab:university1652} and Table~\ref{tab:sues200}, respectively.

\textbf{Results on the University-1652 Dataset}
As shown in Table~\ref{tab:university1652}, \textbf{S$^3$Geo} achieves absolute state-of-the-art performance, outperforming all existing methods across all evaluation metrics. In the Drone $\rightarrow$ Satellite task, our method reaches 96.19\% on R@1 and 96.86\% on AP, effectively surpassing recent strong competitors like CDM-Net \cite{zhou2025cdm} and ECSNet \cite{yang2026integrating}. Similarly, in the Satellite $\rightarrow$ Drone task, \textbf{S$^3$Geo} achieves 96.72\% R@1 and 95.10\% AP, outperforming all competing methods. This consistent superiority, particularly the significant gain in AP, demonstrates that our method is not only accurate in identifying the top-1 match but also robust in retrieving all relevant images. These improvement are likely attributed to our synergistic design: the semantic priors effectively filter out visually similar but contextually incorrect distractions, while the DQP ensures precise local structural alignment.

\textbf{Results on the SUES-200 Dataset.}
The SUES-200 dataset introduces significant challenges due to large scale variations caused by different drone flight altitudes (150m-300m). As shown in Table~\ref{tab:sues200}, \textbf{S$^3$Geo} consistently achieves the best performance across all altitude settings in both retrieval directions. Notably, for the Satellite$\rightarrow$Drone task, our method achieves a 100.00\% R@1 across all altitude levels, indicating highly stable cross-view matching performance regardless of scale variations. For the Drone$\rightarrow$Satellite task, \textbf{S$^3$Geo} also reaches near-saturated performance at higher altitudes (e.g., 100.00\% at 300m), while maintaining strong robustness under the most challenging low-altitude setting (150m), achieving 99.23\% R@1. Compared with strong baselines such as ECSNet \cite{yang2026integrating} and MEAN \cite{chen2025multi}, which exhibit noticeable performance degradation at lower altitudes, our method demonstrates superior robustness to severe scale and viewpoint changes. These results validate the effectiveness of jointly modeling structural correspondences and semantic consistency for CVGL.

\subsection{Cross-Region Transferability}

The ability to generalize across different geographic regions is crucial for CVGL, as real-world deployments often involve unseen environments. To evaluate the cross-region generalization capability of \textbf{S$^3$Geo}, we conduct a zero-shot cross-dataset evaluation. Specifically, the model is trained solely on the University-1652 dataset and directly tested on SUES-200 without any fine-tuning, providing a rigorous assessment of domain transferability. As shown in Table~\ref{table:cross-region}, \textbf{S$^3$Geo} consistently outperforms existing methods in both Drone$\rightarrow$Satellite and Satellite$\rightarrow$Drone tasks, achieving the best R@1 and AP across all altitude settings. Compared with purely visual models, which typically suffer from noticeable performance degradation under domain shifts (e.g., variations in illumination, weather, and scene layout), our method demonstrates stronger robustness in cross-dataset scenarios. These results indicate that the proposed structural-semantic synergistic learning enables the model to capture more transferable representations, allowing it to generalize effectively to unseen environments without requiring target-domain adaptation.

\begin{table}[t] 
  \centering
  \scriptsize 
  \caption{Comparison with state-of-the-art results in cross-region transferability. The best results are highlighted in \textbf{bold}, and the second-best are \underline{underlined}.}
  \label{table:cross-region}
  \renewcommand{\arraystretch}{1.05} 
  \setlength{\tabcolsep}{1.5pt}      
  
  \resizebox{\columnwidth}{!}{ 
  \begin{tabular}{l cc cc cc cc} 
    \toprule
    \multicolumn{9}{c}{\textbf{Drone$\rightarrow$Satellite}} \\ 
    \midrule
    \multirow{2}{*}{\textbf{Method}} &
    \multicolumn{2}{c}{150m} &
    \multicolumn{2}{c}{200m} &
    \multicolumn{2}{c}{250m} &
    \multicolumn{2}{c}{300m} \\ 
    \cmidrule(lr){2-3} \cmidrule(lr){4-5} \cmidrule(lr){6-7} \cmidrule(lr){8-9}
     & R@1 & AP & R@1 & AP & R@1 & AP & R@1 & AP \\ 
    \midrule
    MCCG \cite{shen2023mccg} & 57.62 & 62.80 & 66.83 & 71.60 & 74.25 & 78.35 & 82.55 & 85.27 \\ 
    Sample4Geo \cite{deuser2023sample4geo} & 70.05 & 74.93 & 80.68 & 83.90 & 87.35 & 89.72 & 90.03 & 91.91 \\ 
    CAMP \cite{wu2024camp} & 78.90 & 82.38 & 86.83 & 89.28 & 91.95 & 93.63 & \underline{95.68} & \underline{96.65} \\ 
    DAC \cite{xia2024enhancing} & 76.65 & 80.56 & 86.45 & 89.00 & \underline{92.95} & 94.18 & 94.53 & 95.45 \\ 
    MEAN \cite{chen2025multi} & 81.73 & 87.72 & 89.05 & \underline{91.00} & 92.13 & 93.60 & 94.63 & 95.76 \\
    ECSNet \cite{yang2026integrating} & \underline{82.85} & \underline{88.02} & \underline{89.21} & \underline{91.00} & 92.89 & \underline{94.20} & 94.66 & 95.80 \\
    \midrule
    \rowcolor{gray!10} \textbf{S$^3$Geo (Ours)} & \textbf{91.65} & \textbf{93.46} & \textbf{96.73} & \textbf{97.45} & \textbf{98.40} & \textbf{98.74} & \textbf{99.23} & \textbf{99.36} \\
    \midrule
    \multicolumn{9}{c}{\textbf{Satellite$\rightarrow$Drone}} \\ 
    \midrule
    \multirow{2}{*}{\textbf{Method}} &
    \multicolumn{2}{c}{150m} &
    \multicolumn{2}{c}{200m} &
    \multicolumn{2}{c}{250m} &
    \multicolumn{2}{c}{300m} \\ 
    \cmidrule(lr){2-3} \cmidrule(lr){4-5} \cmidrule(lr){6-7} \cmidrule(lr){8-9}
     & R@1 & AP & R@1 & AP & R@1 & AP & R@1 & AP \\ 
    \midrule
    MCCG \cite{shen2023mccg} & 61.25 & 53.51 & 82.50 & 67.06 & 81.25 & 74.99 & 87.50 & 80.20 \\ 
    Sample4Geo \cite{deuser2023sample4geo} & 83.75 & 73.83 & 91.25 & 83.42 & 93.75 & 89.07 & 93.75 & 90.66 \\ 
    CAMP \cite{wu2024camp} & 87.50 & 78.98 & 95.00 & 87.05 & \underline{95.00} & 91.05 & \underline{96.25} & 93.44 \\ 
    DAC \cite{xia2024enhancing} & 87.50 & 79.87 & 96.25 & 88.98 & \underline{95.00} & 92.81 & \underline{96.25} & 94.00 \\ 
    MEAN \cite{chen2025multi} & 91.25 & 81.50 & 96.25 & 89.55 & \underline{95.00} & 92.36 & \underline{96.25} & 94.32 \\
    ECSNet \cite{yang2026integrating} & \underline{91.28} & \underline{81.52} & \underline{96.28} & \underline{89.83} & \underline{95.00} & \underline{92.82} & \underline{96.25} & \underline{94.36} \\
    \midrule
    \rowcolor{gray!10} \textbf{S$^3$Geo (Ours)} & \textbf{96.25} & \textbf{94.45} & \textbf{97.50} & \textbf{96.71} & \textbf{97.50} & \textbf{97.15} & \textbf{97.50} & \textbf{97.43} \\
    \bottomrule
  \end{tabular}
  }
  \vspace{-0.3cm}
\end{table}

\subsection{Ablation Study}

To validate the core motivation of \textbf{S$^3$Geo}, we conduct hierarchical ablation studies on the University-1652 dataset, including both overall component analysis and internal design evaluation.

\textbf{Overall Component Analysis.}
We first evaluate the effectiveness of the main components, including DQP and SKD. Starting from the global baseline, introducing either DQP or SKD leads to consistent performance improvements, while combining both yields the best result. This demonstrates that structural modeling and semantic enhancement are both essential and complementary for CVGL.

\textbf{Internal Design Analysis.}
We further analyze the internal design of DQP and SKD, as shown in Table~\ref{tab:ablation}. Starting from the global baseline, introducing DQP with naive matching brings moderate improvements, indicating that region-aware feature aggregation is beneficial for cross-view matching. Replacing naive matching with OT-based alignment further improves performance, demonstrating that modeling soft correspondences effectively alleviates spatial misalignment across views.

For semantic knowledge distillation, both feature-level and relation-level distillation independently improve performance over the baseline, suggesting that semantic priors provide complementary information beyond purely visual features. Notably, relation-level distillation yields slightly larger gains, highlighting the importance of preserving similarity structure. Combining both further enhances performance, indicating that feature alignment and relational consistency are mutually beneficial.

Finally, integrating both DQP and SKD achieves the best result, confirming that structural modeling and semantic enhancement are complementary and jointly contribute to more discriminative and robust cross-view representations.

\begin{table}[t]
  \centering
  \small
  \caption{Ablation study of the internal design of DQP and SKD on the University-1652 dataset.}
  \label{tab:ablation}
  \renewcommand{\arraystretch}{1}
  \setlength{\tabcolsep}{3pt} 
  \resizebox{\columnwidth}{!}{ 
  \begin{tabular}{l cc cc}
    \toprule
    \multirow{2}{*}{\textbf{Variant}} & \multicolumn{2}{c}{\textbf{Drone$\rightarrow$Satellite}} & \multicolumn{2}{c}{\textbf{Satellite$\rightarrow$Drone}} \\
    \cmidrule(lr){2-3} \cmidrule(lr){4-5}
     & \textbf{R@1} & \textbf{AP} & \textbf{R@1} & \textbf{AP} \\
    \midrule

    \multicolumn{5}{c}{\textit{(a) Effect of Decoupled Query Pooling and OT}} \\
    \midrule
    Baseline & 95.12 & 96.02 & 96.29 & 93.79 \\
    + DQP (w/o OT) & 95.42 & 96.20 & 96.37 & 94.65 \\
    + DQP (w/ OT) & 95.80 & 96.50 & 96.58 & 94.90 \\

    \midrule
    \multicolumn{5}{c}{\textit{(b) Effect of Semantic Knowledge Distillation}} \\
    \midrule
    Baseline & 95.12 & 96.02 & 96.29 & 93.79 \\
    + SKD (feature only) & 95.63 & 96.38 & 96.52 & 94.80 \\
    + SKD (relation only) & 95.68 & 96.42 & 96.54 & 94.86 \\
    + SKD (full) & 95.86 & 96.62 & 96.64 & 94.92 \\

    \midrule
    \rowcolor{gray!10}
    \textbf{Full Model (Ours)} & \textbf{96.19} & \textbf{96.86} & \textbf{96.72} & \textbf{95.10} \\

    \bottomrule
  \end{tabular}
  } 
\end{table}

\textbf{Backbone and Teacher Robustness.}
As shown in Table~\ref{tab:backbone_teacher}, S$^3$Geo consistently improves the corresponding baseline under both DINOv2 \cite{oquab2023dinov2} and DINOv3 \cite{simeoni2025dinov3}. With DINOv2, it improves R@1 by 2.09 and 1.11 percentage points for Drone$\rightarrow$Satellite and Satellite$\rightarrow$Drone, respectively, confirming that the gains do not solely arise from DINOv3. Moreover, replacing CLIP with SigLIP \cite{zhai2023sigmoid} or RemoteCLIP \cite{liu2024remoteclip} results in R@1 differences below 0.2 percentage points, indicating that SKD is robust to the choice of semantic teacher.

\begin{table}[t]
  \centering
  \small
  \caption{Backbone-controlled and semantic-teacher comparisons on University-1652.}
  \label{tab:backbone_teacher}
  \setlength{\tabcolsep}{3pt}
  \renewcommand{\arraystretch}{1.05}

  \resizebox{\columnwidth}{!}{
  \begin{tabular}{ll cc cc}
    \toprule
    \multirow{2}{*}{\textbf{Setting}} &
    \multirow{2}{*}{\textbf{Variant}} &
    \multicolumn{2}{c}{\textbf{Drone$\rightarrow$Satellite}} &
    \multicolumn{2}{c}{\textbf{Satellite$\rightarrow$Drone}} \\
    \cmidrule(lr){3-4} \cmidrule(lr){5-6}
      & & \textbf{R@1} & \textbf{AP} & \textbf{R@1} & \textbf{AP} \\
    \midrule

    \multicolumn{6}{c}{\textit{(a) Backbone-controlled comparison}} \\
    \midrule
    \multirow{2}{*}{DINOv2}
      & Baseline
      & 93.49 & 94.56 & 95.34 & 91.56 \\
      & \textbf{S$^3$Geo}
      & \underline{95.58} & \underline{96.31}
      & \underline{96.45} & \underline{94.65} \\

    \multirow{2}{*}{DINOv3}
      & Baseline
      & 95.12 & 96.02 & 96.29 & 93.79 \\
      & \textbf{S$^3$Geo}
      & \textbf{96.19} & \textbf{96.86}
      & \textbf{96.72} & \textbf{95.10} \\
    \midrule

    \multicolumn{6}{c}{\textit{(b) Effect of semantic teacher}} \\
    \midrule
    \multicolumn{2}{l}{SigLIP-L/16@384}
      & 96.06 & 96.78 & 96.60 & 95.05 \\
    \multicolumn{2}{l}{RemoteCLIP ViT-L/14}
      & 96.14 & 96.83 & 96.63 & 95.09 \\
    \multicolumn{2}{l}{\textbf{CLIP ViT-L/14}}
      & \textbf{96.19} & \textbf{96.86}
      & \textbf{96.72} & \textbf{95.10} \\

    \bottomrule
  \end{tabular}
  }
  \vspace{-0.3cm}
\end{table}

\textbf{Effect of Query Number.}
We investigate the impact of the number of learnable queries $M$ in the DQP module, as shown in Table~\ref{tab:query_number}. When $M=1$, the operation degenerates to producing a single global representation. Although the multi-head mechanism internally attends to various parts, compressing them into one vector fails to explicitly decouple distinct local geometric structures, resulting in inferior performance. Increasing $M$ improves performance by enabling the model to extract multiple region-aware representations. The performance peaks at $M=4$, suggesting a good balance between representation diversity and redundancy. Further increasing the number of queries brings marginal gains or even slight degradation, likely due to overlapping attention regions and increased redundancy. Therefore, we set $M=4$ in all experiments.

\begin{table}[t]
  \centering
  \small
  \caption{Effect of the number of queries $M$ in the DQP module on the University-1652 dataset.}
  \label{tab:query_number}
  \renewcommand{\arraystretch}{1.1}
  \setlength{\tabcolsep}{6pt}
  \begin{tabular}{c cc cc}
    \toprule
    \multirow{2}{*}{\textbf{Queries $M$}} & \multicolumn{2}{c}{\textbf{Drone$\rightarrow$Satellite}} & \multicolumn{2}{c}{\textbf{Satellite$\rightarrow$Drone}} \\
    \cmidrule(lr){2-3} \cmidrule(lr){4-5}
     & \textbf{R@1} & \textbf{AP} & \textbf{R@1} & \textbf{AP} \\
    \midrule
    1 & 95.90 & 96.66 & 96.43 & 94.95 \\
    2 & 95.94 & 96.67 & 96.45 & 94.98 \\
    4 & \textbf{96.19} & \textbf{96.86} & \textbf{96.72} & \textbf{95.10} \\
    8 & 96.11 & 96.81 & 96.71 & 95.09 \\
    \bottomrule
  \end{tabular}
  \vspace{-0.3cm}
\end{table}

\textbf{Sensitivity to Loss Weights.}
We analyze the sensitivity of the loss weights $\lambda_d$ and $\lambda_s$, which control the contributions of DQP and SKD, respectively, as shown in Table~\ref{tab:lambda}. The best performance is achieved at $\lambda_d=0.5$ and $\lambda_s=0.5$, indicating a balanced contribution between structural modeling and semantic supervision. Notably, varying either weight within a reasonable range leads to only minor performance fluctuations, demonstrating that the proposed framework is not sensitive to the specific choice of hyperparameters. This suggests that both structural and semantic components provide stable and complementary contributions, and the model does not rely on carefully tuned weighting to achieve strong performance. Overall, these results indicate that \textbf{S$^3$Geo} is robust to the selection of loss weights, which is desirable for practical deployment.

\begin{table}[t]
  \centering
  \small
  \caption{Sensitivity analysis of the loss weights $\lambda_d$ (DQP) and $\lambda_s$ (SKD) on the University-1652 dataset.}
  \label{tab:lambda}
  \setlength{\tabcolsep}{6pt}
  \begin{tabular}{c c  cc  cc}
    \toprule
    \multirow{2}{*}{$\lambda_d$} & \multirow{2}{*}{$\lambda_s$} & 
    \multicolumn{2}{c}{\textbf{Drone$\rightarrow$Satellite}} & 
    \multicolumn{2}{c}{\textbf{Satellite$\rightarrow$Drone}} \\
    \cmidrule(lr){3-4} \cmidrule(lr){5-6}
     & & R@1 & AP & R@1 & AP \\
    \midrule
    0.30 & 0.50 & 96.00 & 96.71 & 96.58 & 95.04 \\
    \textbf{0.50} & \textbf{0.50} & \textbf{96.19} & \textbf{96.86} & \textbf{96.72} & \textbf{95.10} \\
    0.70 & 0.50 & 95.94 & 96.66 & 96.43 & 94.80 \\
    0.50 & 0.30 & 95.89 & 96.57 & 96.58 & 95.13 \\
    0.50 & 0.70 & 95.93 & 96.69 & 96.62 & 94.99 \\
    \bottomrule
  \end{tabular}
\end{table}

\subsection{Visualization Analysis}

\begin{figure}[h]
  \centering
  \includegraphics[width=0.48\textwidth]{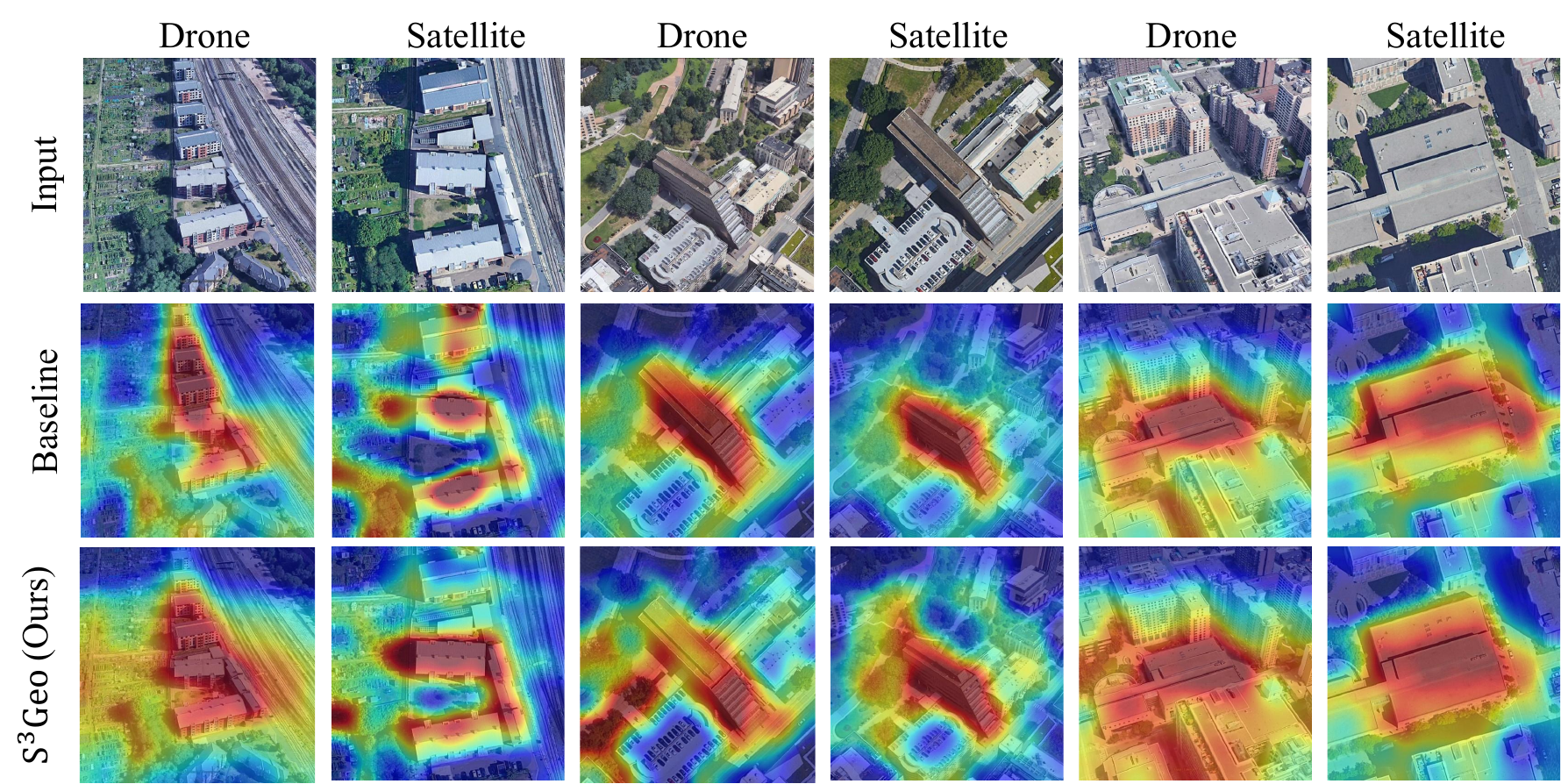} 
  
  \caption{Visualization of heatmaps generated by the backbone network and \textbf{S$^3$Geo} on the University-1652 dataset.}
  \Description{Heatmap comparison between the baseline and S3Geo for three drone and satellite image pairs. S3Geo produces more concentrated attention around relevant structures.}
  \label{fig:heatmap}
\end{figure}

\textbf{Heatmap Visualization.}
We visualize the heatmaps of the baseline and \textbf{S$^3$Geo} to better understand the learned representations. As shown in Figure ~\ref{fig:heatmap}, both models tend to focus on similar salient regions, indicating that the baseline is already capable of capturing coarse structural cues. However, \textbf{S$^3$Geo} produces more concentrated and consistent attention patterns. This suggests that the proposed modules help refine the feature representation by suppressing noise and enhancing meaningful structural information. Although the visual differences are relatively subtle, these improvements lead to more reliable cross-view alignment and better discrimination, which contribute to the overall performance gain.

\section{Conclusion}

We propose \textbf{S$^3$Geo}, a structure-semantic synergistic framework for cross-view geo-localization that explicitly addresses the challenges of structural misalignment and semantic ambiguity across views. By introducing Decoupled Query Pooling with OT-based soft alignment, the model captures fine-grained region-level correspondences, while CLIP-based semantic knowledge distillation enhances semantic consistency and improves discrimination of hard negatives. Extensive experiments on the University-1652 and SUES-200 datasets demonstrate that the proposed method achieves consistent performance gains and strong generalization ability. These results highlight the importance of jointly modeling structural and semantic information for learning robust cross-view representations.

\begin{acks}
  This work was supported by the
  \grantsponsor{yangcheng-scholars}
  {Yangcheng Scholars Research Project}{}
  under Grant No.~\grantnum{yangcheng-scholars}{2024312049},
  the
  \grantsponsor{guangdong-key-lab}
  {Guangdong Provincial Key Lab of Integrated Communication, Sensing and Computation for Ubiquitous Internet of Things}{}
  under Grant No.~\grantnum{guangdong-key-lab}{2023B1212010007},
  and the
  \grantsponsor{cnpc-technology}
  {CNPC Technology Project ``Research on Key Technologies of Artificial Intelligence for Oil and Gas Exploration and Development''}{}
  under Grant No.~\grantnum{cnpc-technology}{2023DJ84}.
\end{acks}

\bibliographystyle{ACM-Reference-Format}
\bibliography{sample-base}










\end{document}